\documentclass{article}

\PassOptionsToPackage{numbers}{natbib}

\usepackage[final]{neurips_2024_ml4ps}

\usepackage[utf8]{inputenc} 
\usepackage[T1]{fontenc}    
\usepackage{hyperref}       
\usepackage{url}            
\usepackage{booktabs}       
\usepackage{amsfonts}       
\usepackage{nicefrac}       
\usepackage{microtype}      
\usepackage{xcolor}         

\usepackage{amsmath}

\usepackage{amssymb}
\usepackage{bm}
\usepackage{cuted}
\usepackage{booktabs}
\usepackage{wrapfig}
\usepackage{xcolor}
\usepackage{graphicx}
\usepackage{multirow}
\usepackage{wrapfig}
\usepackage{caption}
\usepackage{subcaption}
\usepackage{tablefootnote}
\usepackage{comment}

\newcommand{\qcr}[1]{{\fontfamily{cmtt}\selectfont #1}}
\newcommand{\nig}{\text{\qcr{NIG}}}

\newcommand\E{\mathbb{E}}
\newcommand{\Var}{\mathrm{Var}}

\newcommand{\Loss}{\mathcal{L}}
\newcommand{\Lnll}{\Loss^{\scalebox{.7}{{NLL}}}}
\newcommand{\Lreg}{\Loss^{\scalebox{.7}{{R}}}}

\def\enddisplaymath{\]\@ignoretrue}

\hypersetup{
  colorlinks=true,
  citecolor=blue,
  linkcolor=black,
  urlcolor=blue,
  pdfborder={0 0 0}
}

\title{Evidential Deep Learning for Probabilistic Modelling of Extreme Storm Events}

\author{
  Ayush Khot${}^{1,\dagger}$, Xihaier Luo${}^{2}$, Ai Kagawa${}^{2}$, 
  Shinjae Yoo${}^{2}$\\
  ${}^{1}$University of Illinois at Urbana-Champaign, ${}^{2}$Brookhaven National Laboratory\\
  \texttt{\{akhot2\}@illinois.edu, \{xluo, aik, sjyoo\}@bnl.gov} 
}

\begin{document}

\maketitle

\begin{abstract}
   Uncertainty quantification (UQ) methods play an important role in reducing errors in weather forecasting. Conventional approaches in UQ for weather forecasting rely on generating an ensemble of forecasts from physics-based simulations to estimate the uncertainty. However, it is computationally expensive to generate many forecasts to predict real-time extreme weather events. Evidential Deep Learning (EDL) is an uncertainty-aware deep learning approach designed to provide confidence about its predictions using only one forecast. It treats learning as an evidence acquisition process where more evidence is interpreted as increased predictive confidence. 
   We apply EDL to storm forecasting using real-world weather datasets and compare its performance with traditional methods. Our findings indicate that EDL not only reduces computational overhead but also enhances predictive uncertainty. This method opens up novel opportunities in research areas such as climate risk assessment, where quantifying the uncertainty about future climate is crucial. (Github: \url{https://github.com/SULI24/edl-stormcast/})
\end{abstract}

\section{Introduction}


The study of climate and weather holds paramount importance due to its direct impact on natural ecosystems and human societies. Accurate weather predictions can significantly aid in disaster preparedness, agricultural planning, and energy management. Traditionally, numerical weather prediction (NWP) models have been the cornerstone of forecasting. These models simulate the atmosphere and its dynamics based on physical laws. These models, including the state-of-the-art High Resolution Ensemble Forecast (HREF) rainfall nowcasting model used in National Oceanic and Atmospheric Administration (NOAA)~\cite{ravuri2021skilful}, rely on meticulous numerical simulation of physical models. These simulation-based systems fall short in the ability to incorporate signals from newly emerging geophysical observation systems~\cite{goodman2019goes}, or take advantage of the Petabytes-scale Earth observation data~\cite{veillette2020sevir}. In addition, they come with high computational costs and inherent complexities which can limit their scalability and accessibility~\cite{CY47R3-ch5, Leutbecher2019, Zhou2022-nv}.

In recent years, data-driven approaches, particularly deep learning, have emerged as powerful alternatives to traditional NWP models~\cite{reichstein2019deep}. State-of-the-art methods in this domain have demonstrated impressive predictive capabilities by leveraging large datasets of historical weather patterns. Yet, a significant limitation of most existing deep learning approaches is their deterministic nature. Unlike traditional methods, these models typically do not provide measures of uncertainty in their predictions, which is crucial for risk management and decision-making in meteorology~\cite{annurev:/content/journals/10.1146/annurev-statistics-062713-085831, bülte2024uncertaintyquantificationdatadrivenweather}.

Ensemble and Bayesian models address this gap by offering a framework to quantify prediction uncertainty, being especially used in weather forecasting problems~\cite{doi:10.1126/science.1115255, luo2022bayesian, 10.1007/978-981-99-8135-9_29}. Ensemble methods, for instance, generate multiple forecasts to capture a range of possible outcomes, enhancing the reliability of predictions~\cite{10.5555/3295222.3295387}. Bayesian approaches, similarly, provide a probabilistic interpretation by considering the uncertainties in model parameters~\cite{10.5555/3045390.3045502}. Nonetheless, both methods entail substantial computational overheads, primarily due to the need for multiple model evaluations or complex posterior calculations, making them less feasible for real-time forecasting applications. More information on related works are summarized in Appendix~\ref{app:related}.

To bridge this critical gap, we introduce an innovative approach utilizing evidential deep learning~\cite{sensoy2018evidential, amini2020deep}. This method extends the conventional deep learning framework to efficiently quantify uncertainties. Evidence deep learning, based on the concept of evidential theory, estimates the uncertainty directly during the learning process without the need for repetitive model runs or intricate probabilistic sampling. This allows for a significant reduction in computational demands while maintaining the ability to provide calibrated and interpretable uncertainty estimates alongside predictions.

Our empirical results validate that the proposed model not only achieves competitive accuracy in weather forecasting but also produces reliable and well-calibrated uncertainty measurements. These outcomes make a compelling case for the adoption of evidence deep learning in operational settings, where quick and dependable weather forecasts are crucial.





\section{Methods}

\subsection{Problem Statement}
\label{sec:prob}

The primary objective in weather forecasting is the accurate prediction of future atmospheric states based on a series of observed historical data. { \color{black} Let \( x_t \) denote the atmospheric state at time \( t \), encapsulating various meteorological variables such as precipitation (kg\,m\( ^{-2} \)). }
The task is to forecast the future $k$ time step states \( x_{t+1}, x_{t+2}, \ldots, x_{t+k} \) by leveraging the historical sequence of the previous $n$ time steps \( x_{t-n}, x_{t-n+1}, \ldots, x_t \).
The predictive task can be formalized as a function mapping from the domain of past atmospheric data to the domain of future states:
\begin{equation}
    f: \mathcal{X}^{n+1} \rightarrow \mathcal{X}^k
\end{equation}
where \( \mathcal{X} \) represents the set of atmospheric states, \( n \) indicates the number of historical time steps utilized, \( k \) specifies the number of future time steps to be forecasted, and \( f \) is the predictive function.

\subsection{Evidential Deep Learning}
\label{sec:edl}

The central objective of this research is to develop an approximation of the function \( f \) using a sophisticated deep learning model, denoted as \( \hat{f} \). For this purpose, we have selected EarthFormer, a state-of-the-art architecture known for its effectiveness in handling complex spatio-temporal data, as our primary model. However, the inherent challenge with such deterministic models lies in their inability to effectively capture the uncertainty intrinsic to meteorological phenomena, which is crucial given the chaotic nature of weather systems. To address this limitation, we propose to enhance the EarthFormer model by extending its capabilities into the probabilistic domain through the integration of evidential deep learning (EDL). Due to page constraints, the comprehensive details of the EarthFormer are relegated to the Appendix~\ref{app:model}. This section concentrates on the implementation of the EDL framework.

We are given a dataset, $\mathcal{D}$ with $N$ paired training examples, $\mathcal{D} = \{\bm{x}_i, y_i\}_{i=1}^N$. In the deep evidential regression framework, the targets, $y_i$, are parameterized by a Gaussian distribution with an unknown mean and variance \( (\mu, \sigma^2 )\)~\cite{amini2020deep, sensoy2018evidential}. The conjugate prior distribution of \( (\mu, \sigma^2 )\) is then set to the Normal-Inverse-Gamma (\nig{}) distribution~\cite{parisi1988statistical}.
Deep evidential regression alters the model to output the parameters, \( \bm{m}  = (\gamma, \upsilon, \alpha, \beta) \), of the higher-order, evidential \nig{} distribution where $\gamma \in \mathbb{R}$, $\upsilon > 0$, $\alpha > 1$, $\beta > 0$.
. Since \( \bm{m} \) is composed of 4 parameters, the model needs four output neurons for every target parameter. To enforce the constraints on $(v, \alpha, \beta)$, we use a \text{\qcr{softplus}}{} activation (and additional $+1$ added to $\alpha$ since $\alpha > 1$). A linear activation is used for $\gamma \in \mathbb{R}$. Then given a \nig{} distribution, we can then compute the prediction, aleatoric, and epistemic uncertainty as:
\begin{align}
    \underbrace{\E[\mu]=\gamma}_{\text{prediction}}, \qquad \underbrace{\E[\sigma^2]=\tfrac{\beta}{\alpha-1}}_{\text{aleatoric uncertainty}}, \qquad \underbrace{\Var[\mu]=\tfrac{\beta}{\upsilon(\alpha-1)}}_{\text{epistemic uncertainty}}.
\end{align}
To ensure the model learns these parameters, the optimal loss function for the EDL model is composed of two primary components: the negative log likelihood, $\Lnll_i$, and the evidence regularizer, $\Lreg_i$. The authors of Ref.~\cite{amini2020deep} show that by using Bayesian probability theory, $\Lnll_i$ becomes a scaled Student's $t$-distribution parameterized as:
\begin{equation}
\Lnll_i(\bm w) = - \log \text{St}\left(y_i; \gamma, \frac{\beta(1+ \upsilon)}{\upsilon\,\alpha} , 2\alpha\right).
\label{eq:post_pred}
\end{equation}
where $\text{St}\left(y; \mu_\text{St}, \sigma_\text{St}^2, \upsilon_{St}\right)$ is the Student-t distribution evaluated at $y$ with location $\mu_\text{St}$, scale $\sigma_\text{St}^2$, and $\upsilon_{St}$ degrees of freedom. Instead of using the KL Divergence in $\Lreg_i$ like in the classification setting, the authors formulate a novel evidence regularizer as:
\begin{equation}
\Lreg_i(\bm w) = | y_i - \gamma | \cdot (2 \upsilon + \alpha)
\label{eq:post_pred}
\end{equation}
The total loss, $\Loss_i(\bm w)$, is composed of the two loss terms for maximizing and regularizing evidence, scaled by a regularization coefficient, $\lambda$,
\begin{align}
    \Loss_{i}(\bm w) = \Lnll_i(\bm w) + \lambda \, \Lreg_i(\bm w).
\end{align}
where $\lambda > 0$ is a regularization coefficient. Here, $\lambda$ trades off uncertainty inflation with model fit. Setting $\lambda=0$ yields an over-confident estimate while setting $\lambda$ too high results in over-inflation. The authors of Ref.~\cite{sensoy2018evidential} proposed a dynamically scaled choice of $\lambda$ to ensure a gradual increase of $\lambda$ during the training process. This scaling allows the influence of the evidence regularizer to initially be limited, avoiding overly harsh penalties that could lead to model convergence towards an under-confident distribution prematurely. Then, as the model converges, the evidence regularizer becomes more prominent, guiding the model towards a more accurate uncertainty quantification.

\section{Experiments}

\subsection{Experimental Setup}

\textbf{Data} We utilize the Storm EVent ImageRy (SEVIR) dataset~\cite{veillette2020sevir}. Specifically, the task is defined as precipitation nowcasting by predicting the next 12 images,
each representing a 5-minute interval, in the sequence given 13 images, or 65 minutes, additionally detailed in Appendix~\ref{app:data}.

\textbf{Baselines} The proposed EDL is assessed against two state-of-the-art models: ensemble methods~\cite{10.5555/3295222.3295387} and Monte Carlo (MC) Dropout~\cite{10.5555/3045390.3045502}. Both methods require multiple inferences to estimate the uncertainty. As a result, we use 10 different inference passes. Details are available in Appendix~\ref{app:base}.

\subsection{Results}

\textbf{Model Accuracy} All training is done on two NVIDIA H100 GPUs, and all testing is done on 1 NVIDIA A100 GPU. We first calculate the Critical Success Index (CSI), a widely used metric in precipitation nowcasting that evaluates forecast accuracy by comparing correctly predicted events to the total number of predicted or observed events. CSI is usually defined as $\frac{\text{\#Hits}}{\text{\#Hits} + \text{\#Misses} + \text{\#FalseAlarms}}$, where \#Hits (truth=1, pred=1), \#Misses (truth=1, pred=0), and \#FalseAlarms (truth=0, pred=1) are determined after rescaling predictions and ground truth to 0-255 and binarizing at thresholds \([16, 74, 133, 160, 181, 219]\). CSI ranges from 0 to 1, with 1 indicating a perfect forecast. In Table~\ref{table:sevir_csi}, we report CSI values across different thresholds. Results indicate that EDL performs relatively better at lower thresholds, though it lags behind Ensemble and MC Dropout methods. {\color{black} Notably, using initial weights pretrained without EDL significantly boosts EDL performance, especially at higher thresholds like CSI-181 and CSI-160.}

\begin{table}[h]
\vskip -0.3cm
\caption{Prediction performance comparison. Note that P-EDL refers to EDL with pretrained weights optimized using MSE loss.}
\label{table:sevir_csi}
	\begin{center}
	\resizebox{0.90\textwidth}{!}{
	\begin{tabular}{l|cccccc}
	\toprule[1.5pt]
	\multirow{2}{*}{Model} & \multicolumn{6}{|c}{Metrics}\\

 	& $\mathtt{CSI}\mbox{-}219$ $\uparrow$ & $\mathtt{CSI}\mbox{-}181$ $\uparrow$ & $\mathtt{CSI}\mbox{-}160$ $\uparrow$ & $\mathtt{CSI}\mbox{-}133$ $\uparrow$ & $\mathtt{CSI}\mbox{-}74$ $\uparrow$ & $\mathtt{CSI}\mbox{-}16$ $\uparrow$ \\
	\midrule\midrule
	Ensemble & 0.1436 & 0.2613 & 0.3081 & 0.4225 & 0.6947 & 0.7666 \\
	MC Dropout & 0.1436 & 0.2613 & 0.3081 & 0.4225 & 0.6947 & 0.7666 \\
	EDL	&  0.0027 & 0.0726 & 0.1530 & 0.3303 & 0.6676 & 0.7358 \\
   P-EDL & 0.0066	& 0.1215 & 0.2205 & 0.3682 & 0.6416 & 0.7562 \\
	\bottomrule[1.5pt]
	\end{tabular}
	}
	\end{center}
\end{table}

\begin{minipage}{0.65\textwidth}
We also compute the mean squared error (MSE) as a function of prediction lead time. As shown in Figure~\ref{fig:mse_time}, the MSE for all models increases with longer lead times, highlighting the challenge of maintaining accuracy over extended predictions. This trend also suggests that the Ensemble and MC Dropout methods may be preferable when the primary focus is on minimizing prediction error over time. 
\vspace{0.5em}

\textbf{Model Efficiency} We further assess the computational efficiency of each model by measuring average inference time and GFLOPS. As shown in Figure~\ref{fig:efficiency}, EDL consistently outperforms both MC Dropout and Ensemble in terms of GFLOPS and inference time.
\end{minipage}
\hfill
\begin{minipage}{0.3\textwidth}
    \centering
    \includegraphics[width=1.0\textwidth]{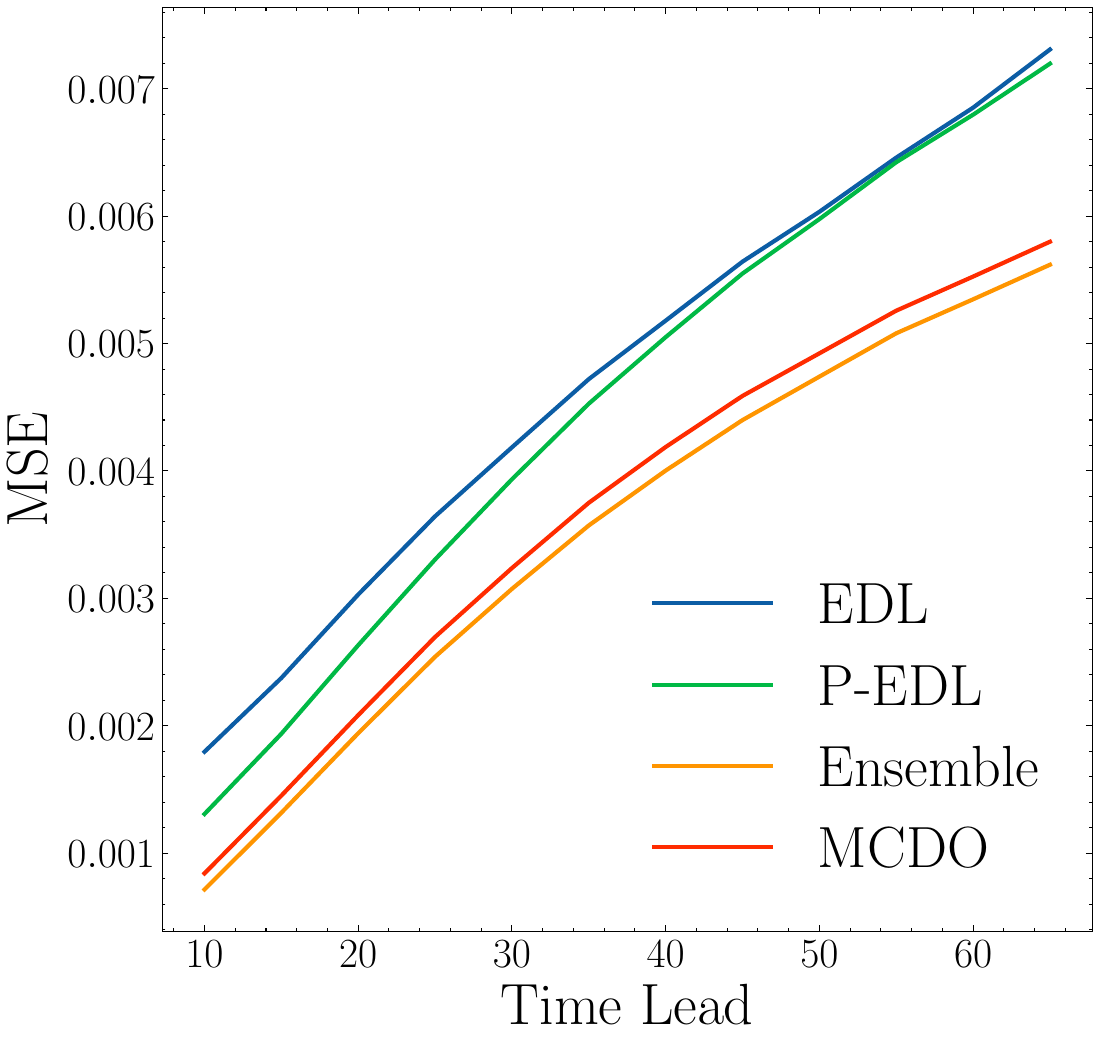}
    \captionof{figure}{\small{Plot of average MSE for varying forecasts time leads.}}
    \label{fig:mse_time}
\end{minipage}

\vspace{-1.3em}
\begin{minipage}{0.37\textwidth}
This advantage arises because EDL conducts uncertainty analysis directly within a single model, whereas MC Dropout and Ensemble require multiple models or particles, necessitating several inferences to estimate uncertainty. Additionally, EDL and MC Dropout have a comparable number of parameters, both significantly fewer than Ensemble, which relies on multiple models. However, MC Dropout's need for repeated runs to generate uncertainty estimates substantially increases its computational cost.
\vspace{0.5em}

\textbf{Uncertainty Diagnosis} In our final analysis, we assess the quality of the estimated uncertainty across different models. To begin, we investigate the relationship between uncertainty and model accuracy by plotting the normalized correlation between uncertainty and MSE at varying forecast time leads (Figure~\ref{fig:uncertainty}). As expected, the correlation generally decreases over time, reflecting the growing difficulty in making accurate predictions as the forecast horizon extends. Nota-

\end{minipage}
\hfill
\begin{minipage}{0.6\textwidth}
    \vspace{2.em}
    \centering
    \includegraphics[width=0.46\textwidth]{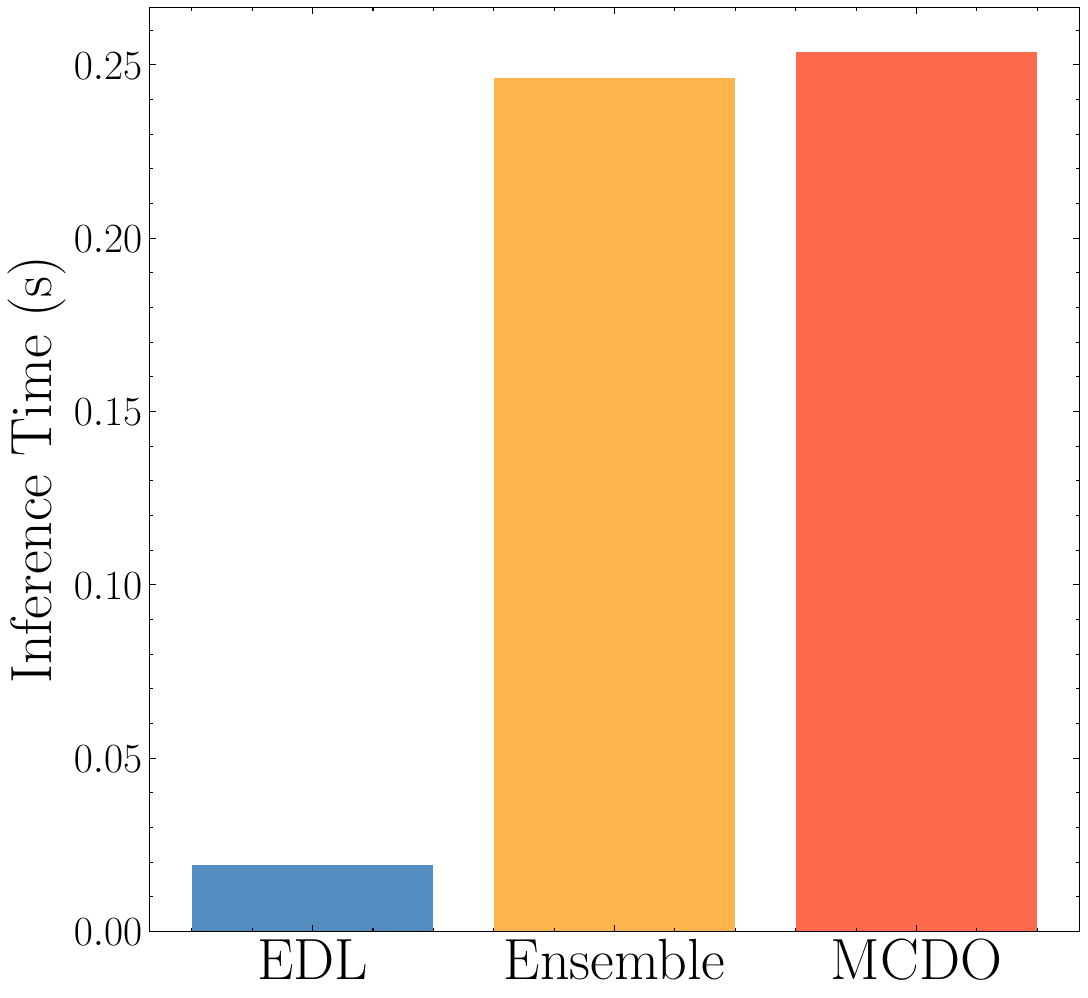} \quad
    \includegraphics[width=0.46\textwidth]{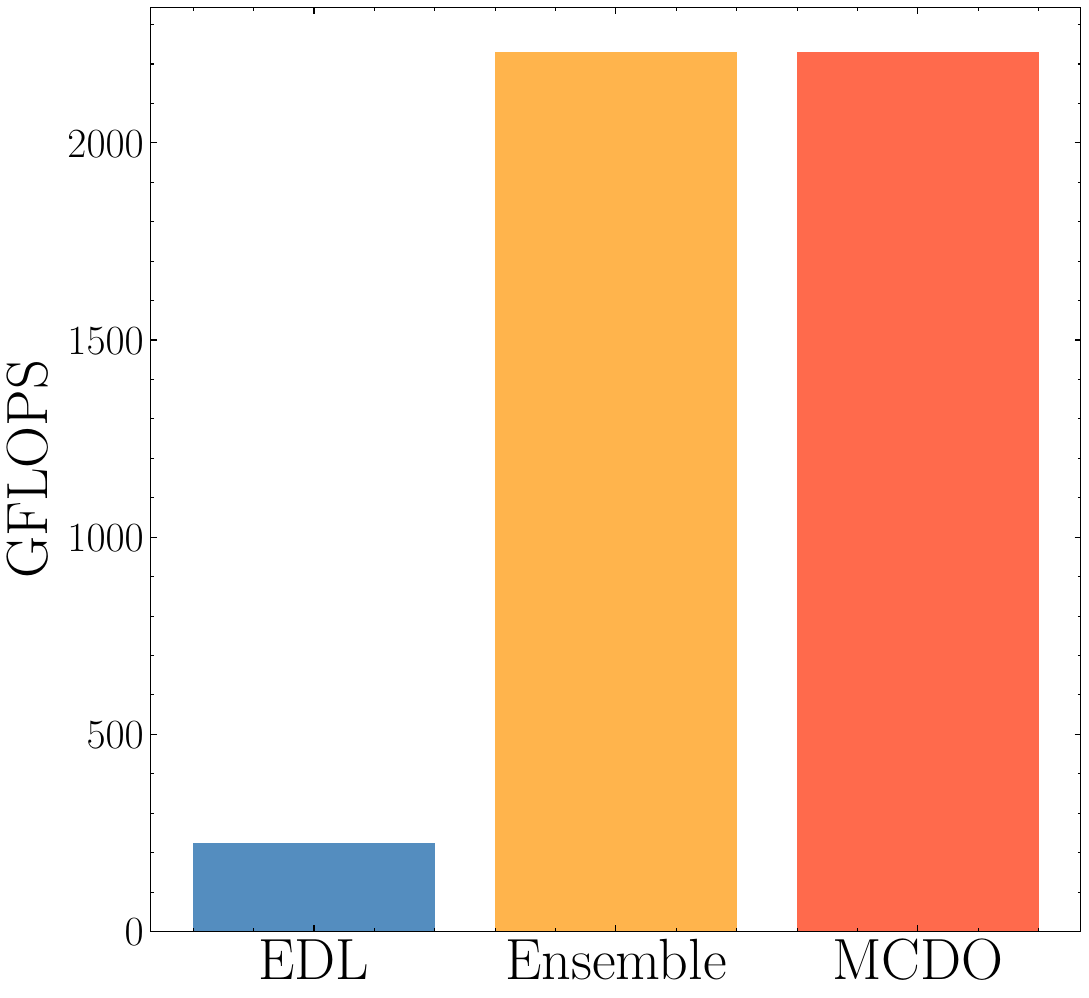}
    \captionof{figure}{\small{Histograms of the  inference time and GFLOPS.}}
    \label{fig:efficiency}
    \vspace{1.em}
    \centering
    \includegraphics[width=0.46\textwidth]{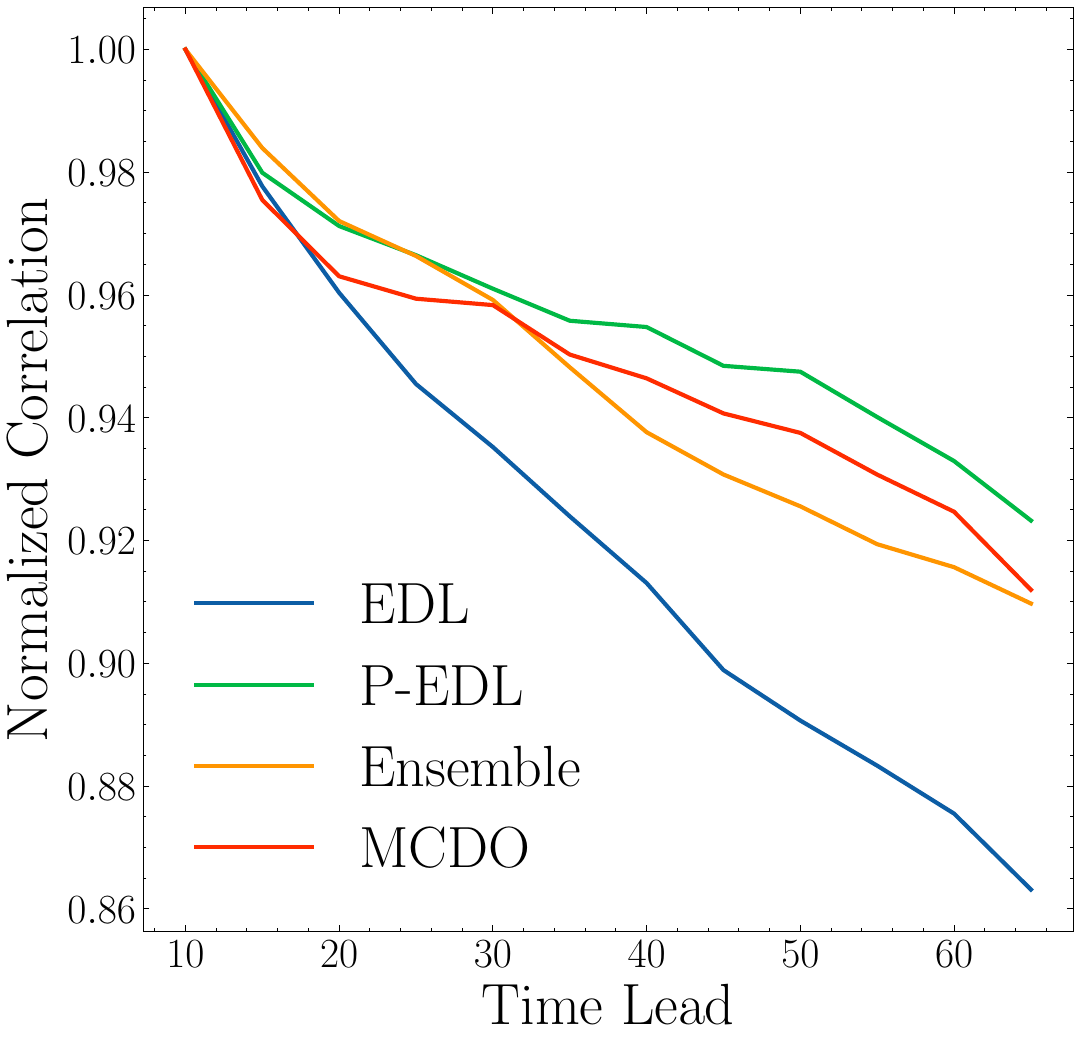} \quad
    \includegraphics[width=0.46\textwidth]{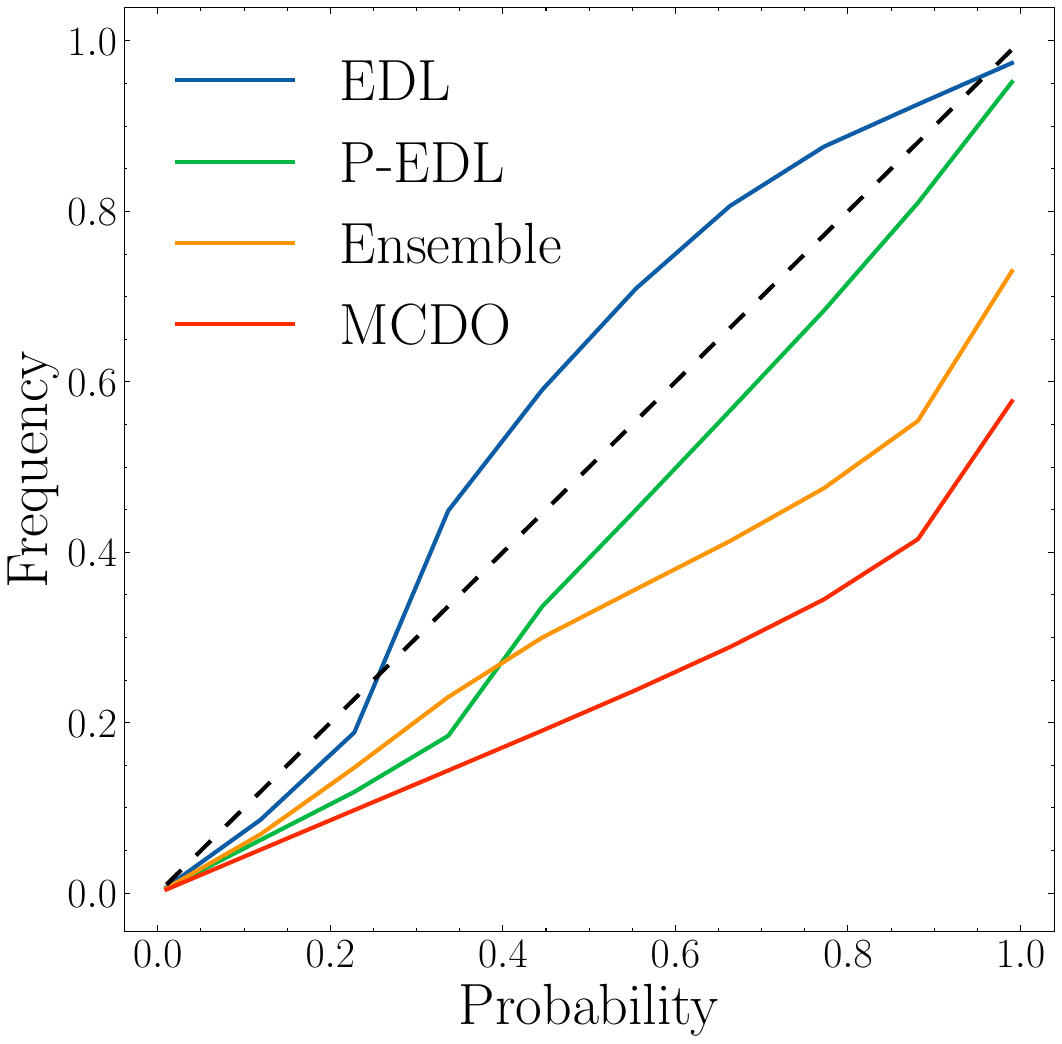}
    \captionof{figure}{\small{Uncertainty analysis results: The left panel displays the normalized correlation between uncertainty and MSE, while the right panel shows the reliability curves. Models closer to the dashed $y=x$ line exhibit well-calibrated uncertainty.}}
    \label{fig:uncertainty}
\end{minipage}

\vspace{-0.3em}
-bly, P-EDL maintains a higher normalized correlation over time, suggesting more stable uncertainty estimates, whereas EDL experiences a sharp decline, indicating a potential loss in calibration with increasing lead time. We also evaluate calibration using a reliability diagram, where a perfectly calibrated model aligns with the dashed \( y = x \) line. EDL demonstrates strong calibration, with P-EDL slightly less accurate. By utilizing pretrained weights, P-EDL prioritizes the accuracy of the prediction rather than the accuracy of the uncertainty. As a result, the accuracy is higher but it has worse uncertainty calibration. In contrast, Ensemble and MC Dropout deviate further from the optimal line, highlighting the superior uncertainty estimation of EDL. Additional uncertainty plots are provided in the Appendix~\ref{app:uncertainty}.

\section{Conclusion}

We propose the use of EDL for storm forecasting as an effective measure of model uncertainty. Although the model has a slightly worse MSE and $\mathtt{CSI}$ values, the use of only one model and one forecast makes it much less time-consuming and computationally expensive than both Ensemble and MC Dropout.  The uncertainty from EDL has also proven to be more well-calibrated than traditional methods. We propose the use of pretrained weights on MSE loss for EDL, P-EDL, to ensure that the MSE loss significantly decreases and it maintains higher normalized correlation with increased leading time. P-EDL ensures similarly well-calibrated uncertainty.
Uncertainty calibration and computational is very important when dealing with reliable, real-time predictions, but it shouldn't come at the expense of accuracy. We will work on subsequent methods to maintain accuracy in EDL.
In the future, EDL could be incorporated to refine the uncertainty estimates across ensemble members, providing a more robust assessment of model predictions by weighting them based on their evidential support. In addition, other data distributions, such as Poisson or a Gamma distribution, could be adopted within EDL.

\begin{ack}
{
\color{black}
This work was supported in part by the U.S. Department of Energy, Office of Science, Office of Workforce Development for Teachers and Scientists (WDTS) under the Science Undergraduate Laboratory Internships (SULI) program. 
This research used resources of the National Energy Research
Scientific Computing Center, a DOE Office of Science User Facility
supported by the Office of Science of the U.S. Department of Energy
under Contract No. DE-AC02-05CH11231 using NERSC award
ASCR-ERCAP0027539.
Xihaier Luo was supported by funding from the Advanced Scientific Computing Research program within the Department of Energy’s Office of Science, under project No. DE-SC0012704. SCIDAC IMPACTS FWP \(\#\) CC127.
Ayush Khot and Ai Kagawa were additionally supported by the concept development funding from the Computing and Data Sciences Directorate at Brookhaven National Laboratory.
The authors acknowledge Nathan Urban for useful discussions, comments and suggestions.
}
\end{ack}

\bibliographystyle{plain}
\bibliography{neurips_2024}


\appendix

\section{Related Works}
\label{app:related}
\paragraph{Deep Learning-based Weather Prediction.}
Traditional DL approaches to weather forecasting involve CNN and RNN. U-Net based architectures that include CNN have been applied to sea ice forecasting~\cite{andersson2021seasonal}, precipitation nowcasting~\cite{veillette2020sevir, Zhang2023}, and cloud cover nowcasting~\cite{FERNANDEZ2021419}.
Shi et al.~\cite{shi2015convolutional} proposed ConvLSTM that leverages CNN and LSTM for precipitation nowcasting. Wang et al.~\cite{wang2022predrnn} proposed PredRNN which extends the predictive capabilities of ConvLSTM by introducing a spatiotemporal memory architecture. 
To better learn long-term high-level relations, Wang et al.~\cite{wang2018eidetic} proposed E3D-LSTM that integrates 3D CNN with LSTM. 
To disentangle PDE dynamics from unknown complementary information, Guen et al. proposed PhyDNet~\cite{guen2020disentangling} which incorporates a new recurrent physical cell to perform PDE-constrained prediction in latent space.  Espeholt et al.~\cite{espeholt2021skillful} proposed MetNet-2, based on ConvLSTM and dilated CNN, that outperforms HREF in precipitation forecasting.
Very recently, there are works that have implemented Transformer for solving weather forecasting problems~\cite{an2024deeplearningprecipitationnowcasting}
Pathak et al.~\cite{pathak2022fourcastnet} proposed the FourCastNet for global weather forecasting, which is based on Adaptive Fourier Neural Operators (AFNO)~\cite{guibas2021adaptive}. Bai et al.~\cite{bai2022rainformer} proposed Rainformer for precipitation nowcasting, which is based on an architecture that combines CNN and Swin-Transformer~\cite{liu2021swin}. Later, Gao et al. proposed the Earthformer~\cite{gao2022earthformer} which implements cuboid attention to reduce computional expense. Pangu-Weather leverages a 3D Vision Transformer that separates the input into cubes to predict in a medium time range scale ($5 - 10$ days). There are also other GNN models like GraphCast~\cite{lam2023learning} and diffusion models like GenCast~\cite{price2024gencastdiffusionbasedensembleforecasting}, but precipitation forecasting has not been their strong point. Diffusion-based models are probabilistic, but sequentially denoising the input over multiple steps can be computationally expensive~\cite{gao2023prediff, Yu2024diffcast, gong2024cascast}.

\paragraph{Probabilistic Modeling}
It is preferred to extend deterministic models to incorporate uncertainty, generating a distribution of possible outcomes rather than a single outcome. There are a variety of popular techniques to incorporate uncertainty:
\textbf{(1) Ensemble Methods:} By combining multiple models or fit one model with diverse hyperparameters, we can get better generalization performance in the final prediction~\cite{10.5555/3295222.3295387}. Popular in NWP models~\cite{doi:10.1126/science.1115255}, ensemble methods has gained traction in deep learning~\cite{ABDAR2021243}. Some examples of its applications include computer vision~\cite{ABDAR2021243}, spatiotermporal forecasting~\cite{https://doi.org/10.1002/env.2553}, mechinal machinery~\cite{Althoff2021-tg}, and precipitation nowcasting~\cite{AMINI2022128197}. However, training multiple models can be time-consuming and computationally intensive.
\textbf{(2) Bayesian Methods:} Instead of using multiple models, Bayesian deep learning (BDL) incorporates Bayesian concepts to quantify uncertainty effectively. BDL models utilize a posterior probability distribution that relies on prior knowledge distribution and the likelihood of the data being utilized~\cite{10.5555/3045390.3045502}. This framework represents model weights as random variables~\cite{ABDULLAH2024e24188}, and the uncertainty can be generated by using the prior and likelihood to sample from the posterior distribution. The most popular techniques are Monte Carlo Dropout~\cite{10.5555/3045390.3045502} and Variational Inference (VI)~\cite{doi:10.1080/01621459.2017.1285773}. These techniques are popular in computer vision~\cite{Song:20, gustafsson2020evaluatingscalablebayesiandeep}, healthcare~\cite{9745083}, and weather forecasting~\cite{luo2022bayesian}. It is also very popular in precipitation nowcasting~\cite{ProbabilisticQuantitativePrecipitationForecastingUsingBayesianModelAveraging, ABayesianApproachforUncertaintyQuantificationofExtremePrecipitationProjectionsIncludingClimateModelInterdependencyandNonstationaryBias, UsingBayesianModelAveragingtoCalibrateForecastEnsembles}. However, it is difficult to scale due to the multiple predictoins needed to approximate uncertainty.
\textbf{(3) Evidential Deep Learning Methods:} Evidential Deep Learning (EDL) is a novel method to quantify uncertainty directly by modeling the evidence supporting different outcomes~\cite{amini2020deep, sensoy2018evidential}. By directly using one model and a single forward pass, EDL offers a comprehensive framework to learn the uncertainty. This technique has been used in healthcare~\cite{Li20213DHM}, computer vision~\cite{Ulmer2021ASO}, and Earth system modeling~\cite{schreck2024evidentialdeeplearningenhancing}. Schreck et al.~\cite{schreck2024evidentialdeeplearningenhancing} also apply EDL in Earth system science applications, but they focus more on Earth system modeling rather than forecasting. Instead, we focus on weather forecasting problems. We examine the computational expense and the uncertainty calibration of EDL compared with baseline methods. 

\section{Model Architecture}
\label{app:model}

The Earthformer model employs a sequence of atmospheric state vectors as its input, representing the dynamic meteorological conditions over time:

\begin{equation}
\mathcal{X}^{n+1} = [x_{t-n}, \ldots, x_t]
\end{equation}

Each vector \( x \) in the sequence encapsulates essential atmospheric variables such as temperature, pressure, and humidity, corresponding to specific timestamps.

\textbf{Input Embedding}: Each input vector \( x_t \) undergoes a transformation into a higher-dimensional feature space to facilitate more complex interactions and learning. This transformation is realized through an embedding layer defined by:

\begin{equation}
\text{Emb}(x_t) = W_e x_t + b_e
\end{equation}

where \( W_e \) and \( b_e \) represent the embedding weight matrix and bias vector, respectively, both of which are parameters learned during training.

\textbf{Positional Encoding}: To incorporate the temporal sequence information essential for forecasting, positional encodings are added to the input embeddings. These encodings are computed using trigonometric functions:

\begin{equation}
PE_{(pos, 2i)} = \sin\left(\frac{pos}{10000^{2i/d_{\text{model}}}}\right), \quad PE_{(pos, 2i+1)} = \cos\left(\frac{pos}{10000^{2i/d_{\text{model}}}}\right)
\end{equation}

where \( pos \) indicates the position in the sequence and \( i \) denotes the dimension index. The embedded inputs are then modified as:

\begin{equation}
H_0 = \text{Emb}(X) + PE
\end{equation}

\textbf{Cuboid Attention}: EarthFormer is a hierarchial Transformer encoder-decoder based on Cuboid Attention. Traditional Transformers compute attention scores using a query (Q), key (K), and value (V) framework, commonly expressed as:

\begin{equation}
\text{Attention}(Q, K, V) = \text{softmax}\left(\frac{QK^T}{\sqrt{d_k}}\right) V
\end{equation}

In this equation, \(Q\), \(K\), and \(V\) are matrices derived from the input data, where \(d_k\) represents the dimensionality of the keys, facilitating the scaling of dot products.

Cuboid attention modifies this framework to address the complexities of 3D data, focusing on the spatial, depth, and temporal dimensions. It is computed by reshaping the embeded inputs \(H_0\) to emphasize its three-dimensional structure:

\begin{equation}
Q, K, V = \text{reshape}(H_0), \quad \text{to shape } (D, H, W, d_k)
\end{equation}

The attention scores are then calculated across blocks or cuboids within the data:

\begin{equation}
\text{Attention}(Q, K, V) = \text{softmax}\left(\frac{\text{reshape}(Q) \cdot \text{reshape}(K)^T}{\sqrt{d_k}}\right) \cdot \text{reshape}(V)
\end{equation}

This reshaping and attention computation enables the model to focus on interactions not just within each plane, but across all three dimensions. To synthesize the information from different dimensions, the attention outputs are combined:

\begin{equation}
H = \frac{1}{3} (\text{Attention}_D(Q, K, V) + \text{Attention}_H(Q, K, V) + \text{Attention}_W(Q, K, V))
\end{equation}

Where \(\text{Attention}_D\), \(\text{Attention}_H\), and \(\text{Attention}_W\) are the attention computations performed along the depth, height, and width dimensions, respectively.

\textbf{Layer Normalization and Feed-Forward Network}: After attention processing, the output undergoes layer normalization to stabilize the training process, followed by a feed-forward network:

\begin{equation}
H' = \text{LayerNorm}(H + \text{Attention}(H)), \quad H'' = \text{LayerNorm}(H' + \text{FFN}(H'))
\end{equation}

The feed-forward network (FFN) consists of two linear transformations with a nonlinear activation function in between, enhancing the model's ability to capture non-linear relationships.

\textbf{Output}: The output from the final Transformer block \( H'' \) is decoded to predict future atmospheric states, which are essential for accurate weather forecasting:

\begin{equation}
\mathcal{X}^k = [x_{t+1}, \ldots, x_{t+k}] = \text{Decoder}(H'')
\end{equation}

where \( \mathcal{X}^k \) denotes the predicted future states, providing crucial insights into upcoming weather conditions.

\section{Additional Details on Experiments}
\label{app:exp}

\subsection{Data}
\label{app:data}
\begin{figure}[h]
    \centering
    \includegraphics[width=0.9\textwidth]{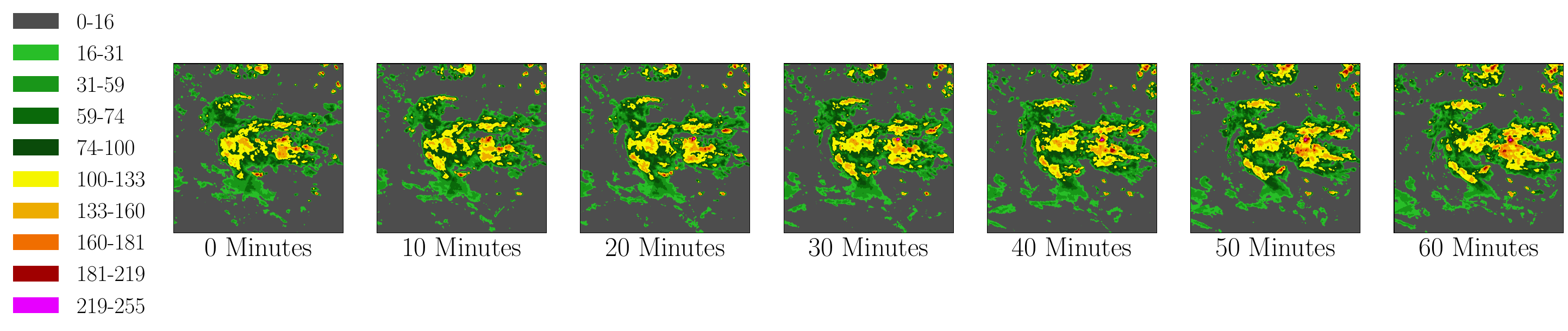}
    \caption{Example Vertically Integrated Liquid (VIL) observation sequence from the Storm EVent ImageRy (SEVIR) dataset. The observation intensity is mapped to pixel value of the range 0-255. The larger value indicates the higher precipitation intensity.}
    \label{fig:sevir_example}
\end{figure}

We utilize the Storm EVent ImageRy (SEVIR) dataset~\cite{veillette2020sevir}, a widely used benchmark for meterological applications. This spatiotemporally aligned dataset contains over 10,000 weather events, each consisting of 384 km $\times$ 384 km image sequences that span over 4 hours. Images in SEVIR were sampled and aligned across five different data types: three channels (C02, C09, C13) from the GOES-16 advanced baseline imager, NEXRAD vertically integrated liquid mosaics, and GOES-16 Geostationary Lightning Mapper  (GLM) flashes. This dataset supports a variety of deep learning research in meteorological applications like precipitation nowcasting, synethic radar generatino, front detection, and more. We use SEVIR for precipitation nowcasting by predicting the next 12 images, or 60 minutes, in the sequence given 13 images, or 65 minutes. Like the authors of Ref.~\cite{gao2022earthformer}, we normalized the data to the range $[0,1]$. Figure~\ref{fig:sevir_example} gives an example of the SEVIR VIL frame sequences.

\subsection{Baselines}
\label{app:base}
Traditionally, ensemble methods~\cite{10.5555/3295222.3295387} and Monte Carlo (MC) Dropout~\cite{10.5555/3045390.3045502} have been popular techniques for estimating epistemic uncertainty in neural networks. Ensemble methods involve training multiple models independently and using the variance of the outputs to evaluate uncertainty. MC Dropout utilizes dropout layers both during training and inference to stimulate the effect of Bayesian inference, thus providing a stochastic basis for uncertainty estimation. Both methods are computationally intensive as they require multiple inferences to estimate the uncertainty, reflecting a significant trade-off between computational efficiency and accuracy. This makes them less prone to be used in real-time due to the large memory and computational expense. For our purposes, we use 10 different inference passes for both Ensemble and MC Dropout.

\section{Uncertainty Plots}
\label{app:uncertainty}

We plot the uncertainties for both the EDL and P-EDL models below using test data. We display the target and output values as well as the Root Mean Square Error (RMSE) and epistemic uncertainty for visualization purposes. We visualize these values over forecasts times from 10 to 40 minutes in the future. The uncertainty tends to get less detailed as the forecast time increases, and the errors also tend to increase. Most of the uncertainties are near the edges of the storms, and they are often concentrated near high VIL values, which the model is insufficient at predicting.

\begin{figure}[h]
    \centering
    \hspace*{-2.0cm}
    \includegraphics[width=1.1\textwidth]{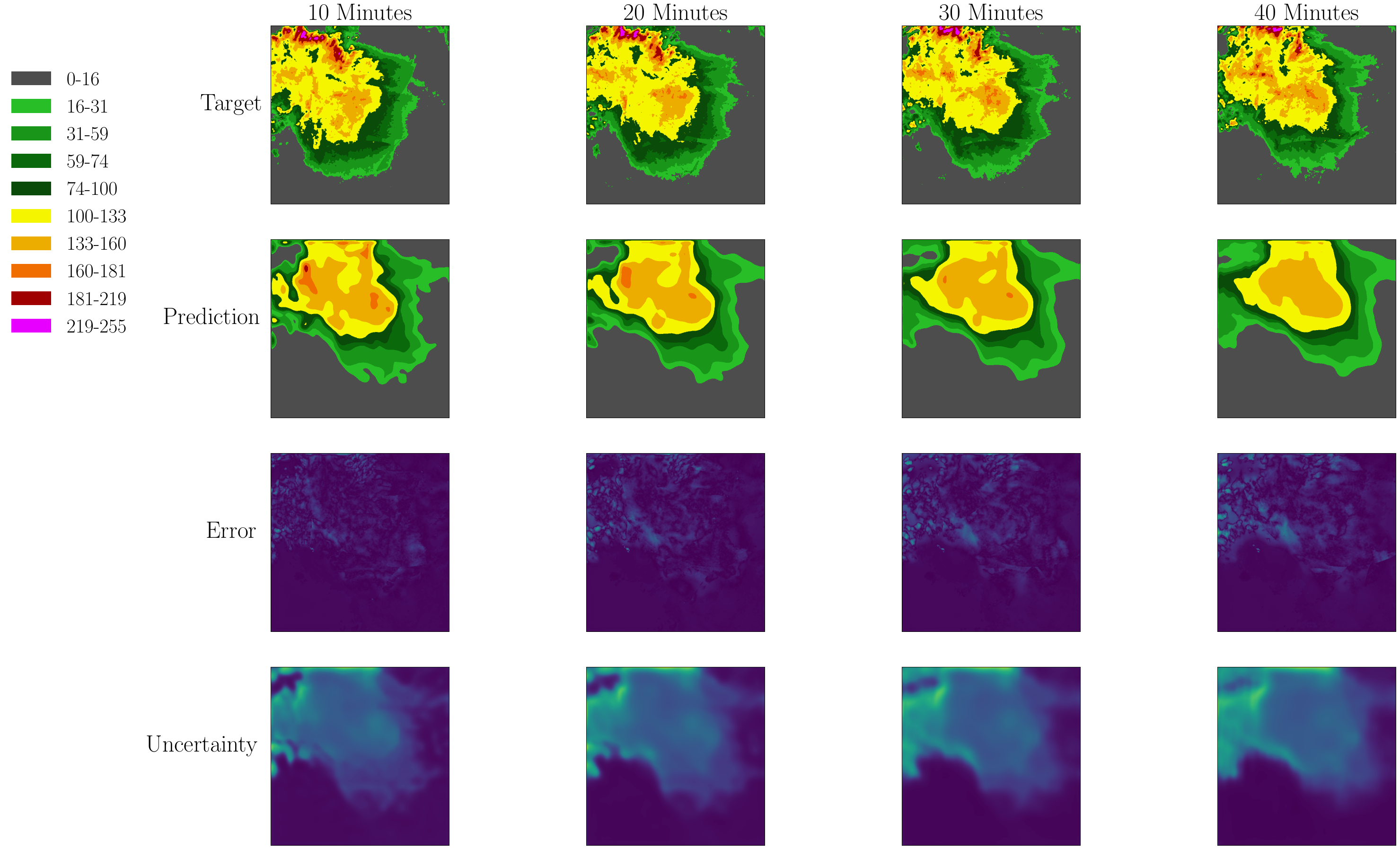}
    \caption{Plot of error and uncertainties for EDL model}
\end{figure}

\begin{figure}[h]
    \centering
    \hspace*{-2.0cm}
    \includegraphics[width=1.1\textwidth]{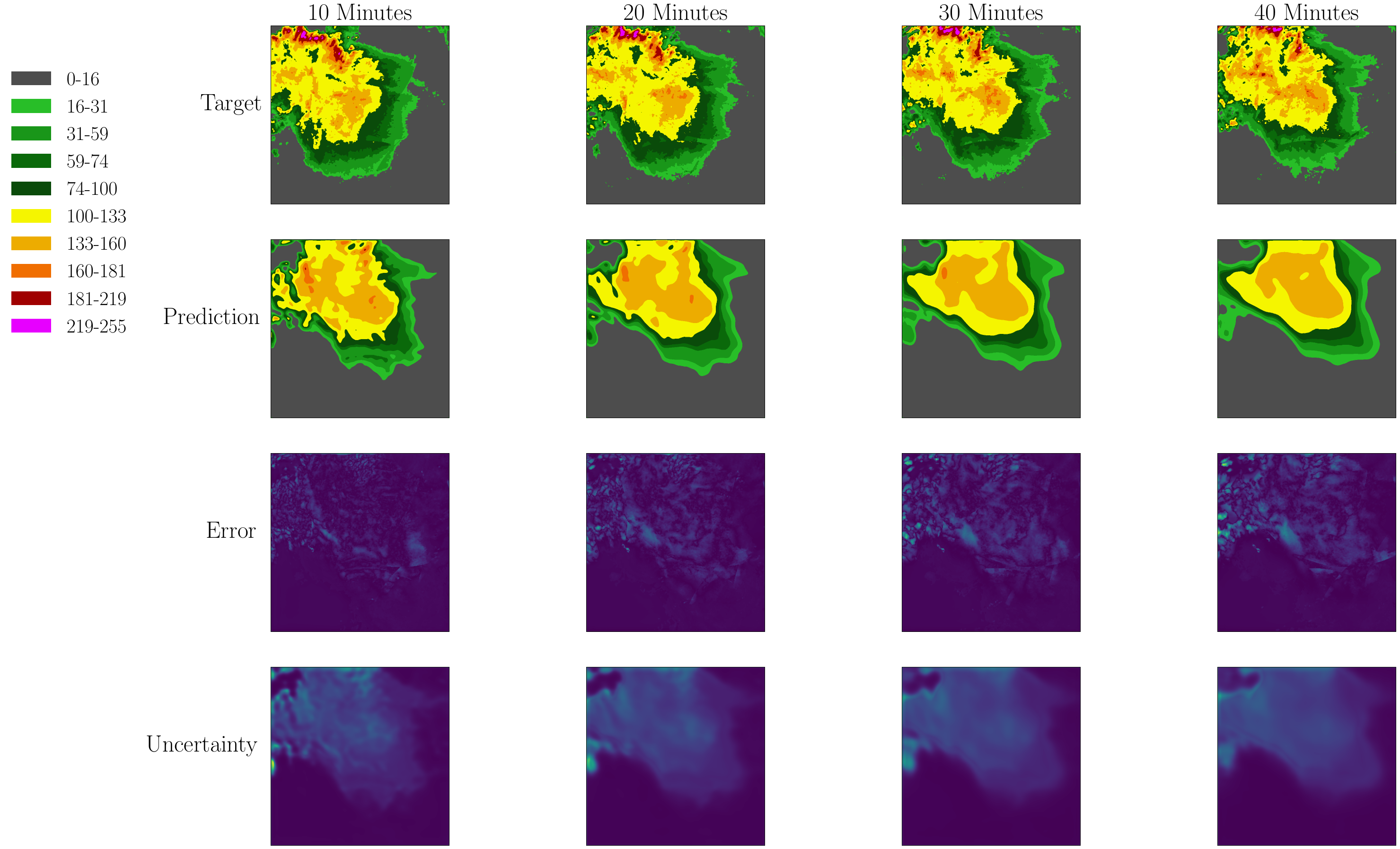}
    \caption{Plot of error and uncertainties for P-EDL model}
\end{figure}

\begin{figure}[h]
    \centering
    \hspace*{-2.0cm}
    \includegraphics[width=1.1\textwidth]{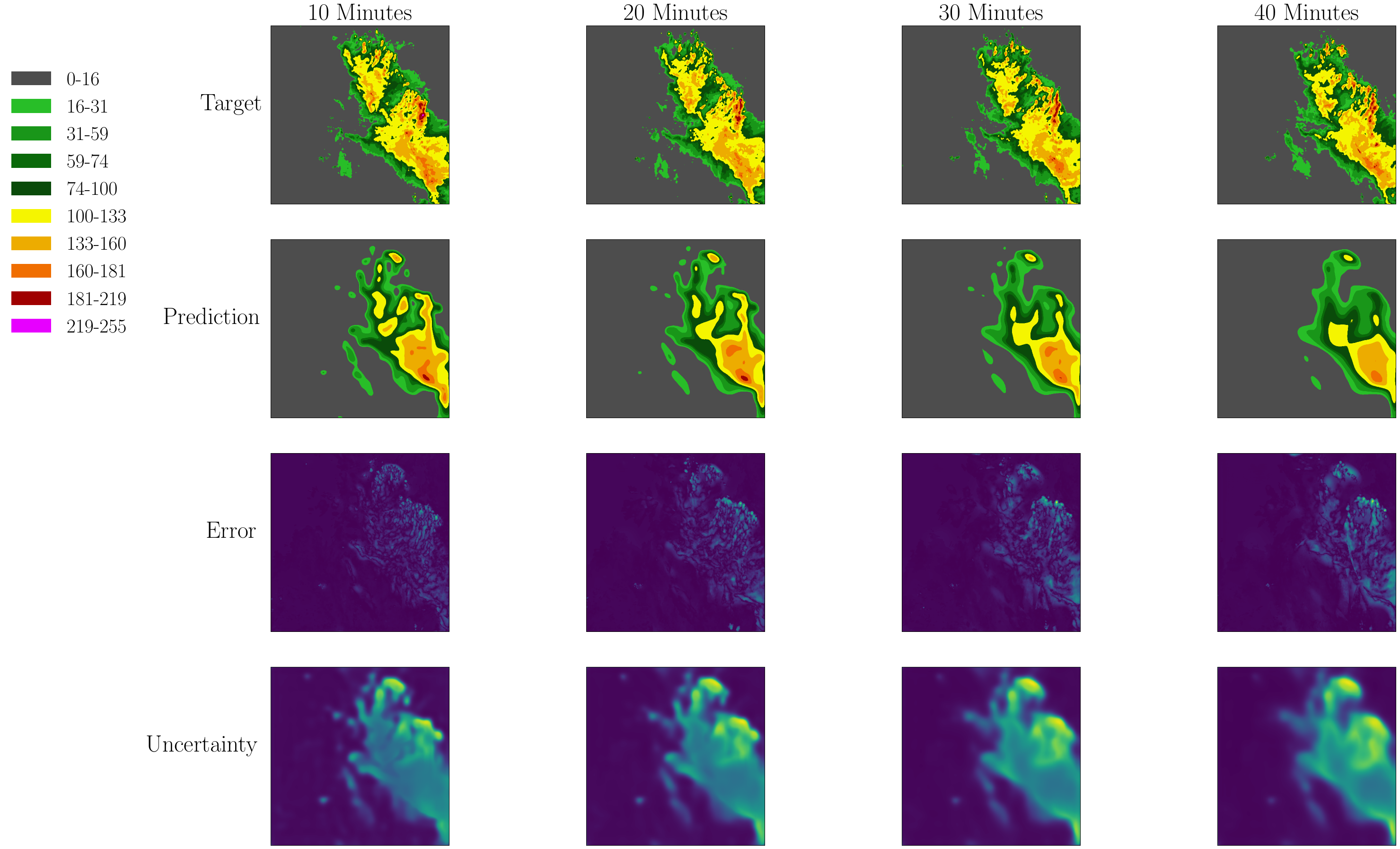}
    \caption{Plot of error and uncertainties for EDL model}
\end{figure}

\begin{figure}[h]
    \centering
    \hspace*{-2.0cm}
    \includegraphics[width=1.1\textwidth]{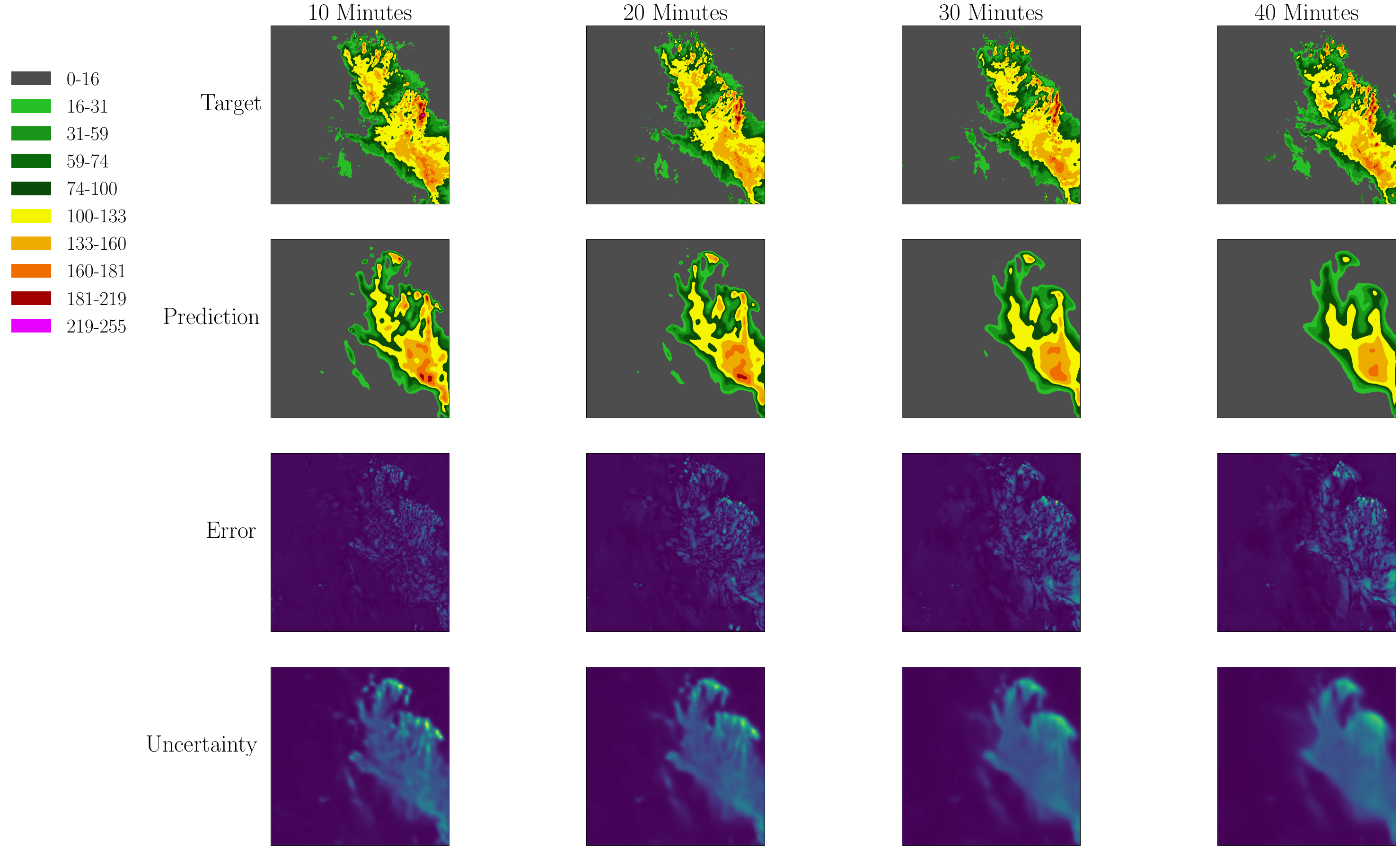}
    \caption{Plot of error and uncertainties for P-EDL model}
\end{figure}

\begin{figure}[h]
    \centering
    \hspace*{-2.0cm}
    \includegraphics[width=1.1\textwidth]{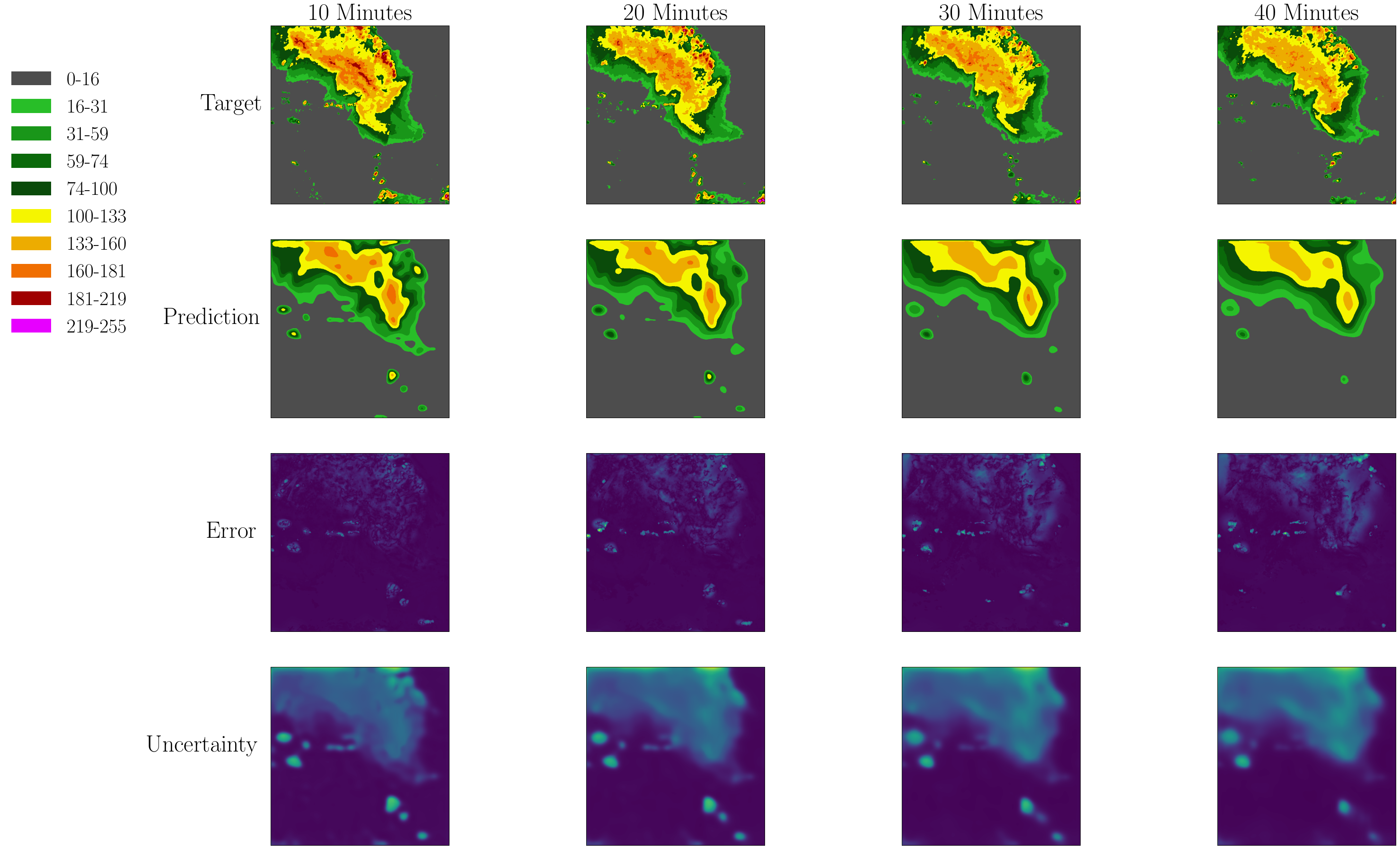}
    \caption{Plot of error and uncertainties for EDL model}
\end{figure}

\begin{figure}[h]
    \centering
    \hspace*{-2.0cm}
    \includegraphics[width=1.1\textwidth]{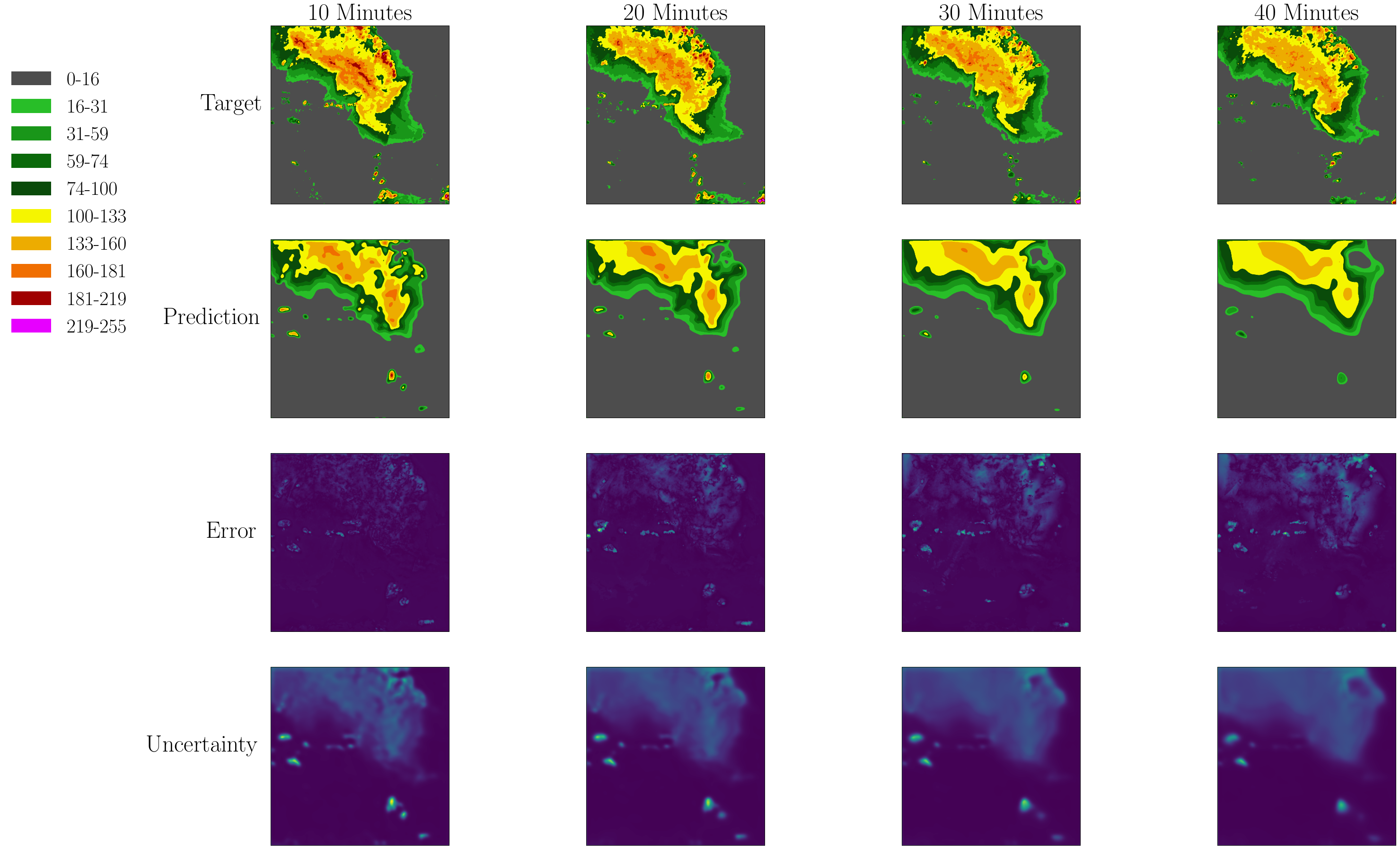}
    \caption{Plot of error and uncertainties for P-EDL model}
\end{figure}

\end{document}